%% file: root.tex
\documentclass[letterpaper, 10 pt, conference]{ieeeconf}  

\IEEEoverridecommandlockouts                              

\title{\LARGE \bf
RGS: Reflection-aware Gaussian Splatting via\\ Learning Geometry Continuity for Reflective Objects
}

\author{
  Xiaobiao Du \textsuperscript{1,3} \space \space \space \space
  Yida Wang  \textsuperscript{3} \space \space \space \space
  Cheng Bi \textsuperscript{3} \space \space \space \space
  Kun Zhan \textsuperscript{3} \space \space \space \space
  Xin Yu \textsuperscript{2}\thanks{Corresponding author: xin.yu@adelaide.edu.au}  \\
  \\
  \textsuperscript{1} University of Technology Sydney   \space \space
    \textsuperscript{2} Adelaide University  \space \space
    \textsuperscript{3} Li Auto Inc. \space \space
}

\usepackage{hyperref}       
\usepackage{url}            
\usepackage{booktabs}       
\usepackage{amsfonts}       
\usepackage{nicefrac}       
\usepackage{microtype}      
\usepackage{xcolor}     

\usepackage{subcaption}
\usepackage{adjustbox}
\usepackage{hhline}
\usepackage{colortbl}
\usepackage{tabularray}
\usepackage{multirow}
\usepackage{physics}
\usepackage{caption}
\usepackage[linesnumbered,ruled]{algorithm2e}

\usepackage{algpseudocode}

\begin{document}

\maketitle
\thispagestyle{empty}
\pagestyle{empty}


\input{sec/0_abstract}

\input{sec/1_intro}

\input{sec/2_related}

\input{sec/3_method}

\input{sec/4_exp}

\input{sec/5_conclude}

\noindent\textbf{Acknowledgments}{
This research is funded in part by ARC-Discovery grant (DP220100800 to XY) and ARC-DECRA grant (DE230100477 to XY). We thank all anonymous reviewers and ACs for their constructive suggestions.}







\bibliography{IEEEfull}
\bibliographystyle{ieeetr}

\end{document}

%% file: sec/0_abstract.tex
\begin{figure*}[h]
    \centering
     \includegraphics[width=\linewidth]{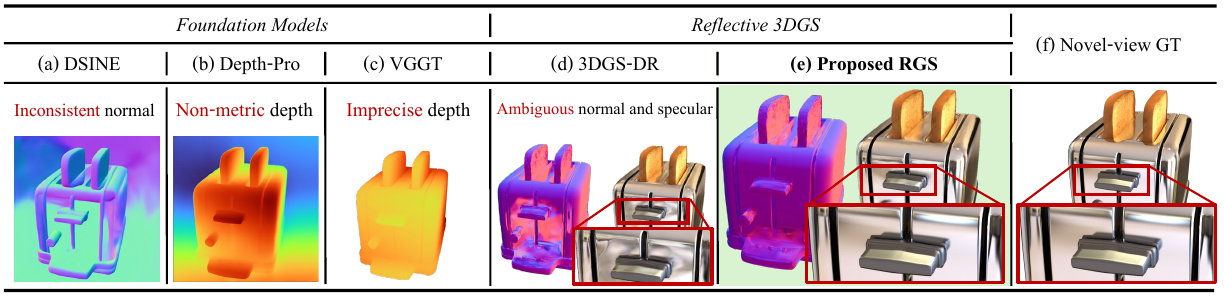} 
    \captionof{figure}{
     \textbf{Our motivation and proposed method.} 
     Existing state-of-the-art reflective 3DGS approaches, \textit{e.g.} 3DGS-DR~\cite{3dgsdr} in (d), struggle to precisely reconstruct the reflective surface, leading to unsatisfactory specular novel view synthesis.
     Traditional foundation models~\cite{bae2024dsine, depthpro} cannot estimate the correct normal in (a) and depth in (b) due to view-dependent ambiguity. We propose to use VGGT in (c) with cross-view geometry constraints for accurate reflective surface supervision.  Thus, our proposed RGS in (e) is a simple yet effective method to regularize the specular surface for reflective objects. Our RGS achieves high-quality novel-view synthesis compared to its ground truth in (f).
    }
    \label{fig:teaser}
    \vspace{-5mm}
\end{figure*}

\begin{abstract}
Gaussian Splatting has significantly improved the quality of novel view synthesis with explicit Gaussian representation.
However, we observed that existing 3D Gaussian Splatting methods (3DGS) often suffer from surface collapse issues on reflective regions, and thus produce inferior geometry and low-quality specular.
In this work, we propose a physically-based deferred rendering framework, named Reflection-aware Gaussian Splatting (RGS), that can accurately model specular regions and improve novel view synthesis performance. 
Specifically, we found that a powerful 3D foundation model can provide a strong 3D geometric prior to foster correct geometric modeling.
Based on this, we propose a cross-view shape consistency regularization to regularize the geometry surface with the large model prior and cross-view constraints.
In this manner, our RGS can produce smoother geometric surfaces on reflective regions while reducing geometric hollows. 
To further improve rendering results on reflective regions, we present a reflection-aware densification strategy that is designed to capture specular variations across various views. With this strategy, our RGS is able to render novel views of objects in higher quality.
Extensive experiments demonstrate our method consistently renders high-quality reflective objects, achieving state-of-the-art performance. \href{https://xiaobiaodu.github.io/reflectivegs/}{\textcolor{magenta}{Project Page: https://xiaobiaodu.github.io/reflectivegs/}}

\end{abstract}

%% file: sec/1_intro.tex
\section{Introduction}
Multi-view stereo (MVS) reconstruction toward 3D scenes and objects has benefited from advancements in volume rendering. In particular, neural radiance field~\cite{mildenhall2021nerf} (NeRF) and neural surface~\cite{wang2021neus, wang2022neus2, wang2022hf} (NeuS) employ multi-layer perceptrons (MLP) to implicitly represent the 3D space with high-frequency details. 
Subsequently, Gaussian Splatting 
(3DGS)~\cite{kerbl2023gaussiansplatting} and its 2D variants~\cite{2dgs}, explicitly represent the scene with learned Gaussian primitives, demonstrating real-time synthesis speed while the high-fidelity rendering~\cite{Yu2024MipSplatting} is guaranteed.
It is demonstrated that high-quality 3D reconstruction can help robotic downstream tasks, such as robotic grasping~\cite{iwase2025zerograsp}, robotic manipulation~\cite{cui2025cl3r}, SLAM~\cite{matsuki2024gaussian}, and self-driving~\cite{du2024dreamcar, du20243drealcar}.
However, reflective object novel view synthesis is very challenging because reflection on a 3D surface is view-dependent, violating cross-view consistency.

Specifically, we found three critical challenges for traditional methods to synthesize high-quality strong-specular objects.
(1) We found that the previous methods, such as 3DGS-DR~\cite{3dgsdr}, tend to suffer from collapsed geometry on reflective surfaces due to overfitting and specular ambiguity as illustrated in Fig.~\ref{fig:teaser} (d). Consequently, the quality of novel view synthesis is inferior; 
(2) Traditional vision foundation models, such as DSINE~\cite{bae2024dsine}, Depth Pro~\cite{depthpro}, and DepthAnything~\cite{depthanything}, fail to estimate normal and depth maps in highly reflective areas, as shown in Fig.~\ref{fig:teaser} (a) and (b). Thus, it is very difficult to obtain accurate geometry in those areas, and this estimated inaccurate geometry cannot assist 3D Gaussians for correct specular mapping;
(3) Although existing volume rendering methods~\cite{refnerf, refneus} can model environment light, they often struggle to capture strong specular reflections, leading to suboptimal rendering quality in reflective regions of novel views.
Therefore, these methods fail to reconstruct high-quality reflective objects when the object is illuminated under strong light.

To solve the aforementioned challenges, we propose a physically-based deferred rendering framework, Reflection-aware Gaussian Splatting, named RGS, with three key components.
(1) Since we found previous methods that tend to produce geometric hollows, we propose physically-based shading and BRDF modeling within deferred rendering. In this fashion, we can reconstruct geometry at a coarse level;
(2) To reconstruct precise geometry, we introduce VGGT~\cite{wang2025vggt}, a recent powerful 3D foundation model that can predict accurate depth information as shown in  Fig.~\ref{fig:teaser} (c).
It can provide cross-view geometric consistency constraints for accurate geometry reconstruction.
In this way, we can further refine the geometry of objects at a fine level;
(3) To better decompose and learn specular information, we propose a reflection-guided densification strategy, enhancing the awareness of 3D Gaussians toward specular. 
With these enhanced 3D Gaussians, the specular effect can be more easily reconstructed, leading to more accurate novel view synthesis results.
In this way, our RGS achieves high-quality geometry continuity and novel view synthesis for reflective objects as shown in Fig.~\ref{fig:teaser} (e).

We conduct extensive experiments to qualitatively and quantitatively validate the efficacy of our proposed RGS. Notably, we visualize the surface geometry, as the normal map, demonstrating that our approach can reliably reconstruct precise and complete surface geometry without structural collapse. This enhancement in terms of geometric accuracy significantly improves the specular novel view synthesis (NVS) for reflective objects. Our method consistently outperforms previous approaches, achieving state-of-the-art results, particularly for reflective objects. In summary, our contributions can be summarized as follows:

$\bullet$ We propose Reflection-aware Gaussian Splatting (RGS) with physically-based shading for reflective object NVS. RGS integrates a cross-view shape regularization strategy, providing cross-view consistency constraints and the 3D foundation model prior to foster accurate specular shading.

$\bullet$ We also propose a reflection-guided densification strategy to enhance the 3D Gaussian primitives that contribute significantly to the rendering of specular, encouraging the learning of specular patterns.

$\bullet$  Extensive experiments in various datasets demonstrate the effectiveness of our method in reconstructing more accurate specular for high-quality novel views synthesis.

%% file: sec/2_related.tex
\section{Related Work}

\textbf{Volume Rendering.}
Unlike explicit 3D representations~\cite{wang2022nvc}, pioneering neural approaches such as NeRF~\cite{f2nerf} and NeuS~\cite{refneus, wang2025hineus} leverage MLPs with volumetric representations to model scenes implicitly. These methods synthesize high-quality novel views without explicit surface or illumination modeling. 
Subsequently, more and more works are proposed to improve the efficiency and quality of volume rendering.
Mip-NeRF 360~\cite{mip-nerf360} reparameterizes the scene for a more compact representation. 
Mip-NeRF~\cite{mip-nerf} and Zip-NeRF~\cite{zip-nerf} design a point sampling strategy for faster ray marching.
In addition, Depth-NeRF~\cite{depthnerf} and Monosdf~\cite{yu2022monosdf} introduce regularization terms to constrain the scene representation, leading to more accurate real-world geometry. 
Although their reconstruction performance is satisfying, their rendering speed is time-consuming due to the numerous ray sampling processes. 
To address this problem, multiple works have been proposed for the acceleration of the rendering process with novel scene representation. NSVF~\cite{liu2020neural} proposes a neural sparse voxel field to replace MLP and improve the rendering efficiency.
Instant-NGP~\cite{instant-ngp} employs hash tables to speed up the rendering process and improve the reconstruction quality.  TensoRF~\cite{TensoRF} reparameterizes the 4D scene tensor into a more compact low-rank representation with vector-matrics decomposition. Recently, Gaussian Splatting ~\cite{3dgsdr}, a Gaussian-based explicit representation method, is proposed and achieves faster rendering speed with better novel view synthesis results.
MVGS~\cite{du2024mvgs} proposes a multi-view training strategy to improve 3D reconstruction quality and novel view synthesis results.
However, these methods are struggling to deal with challenging reflective objects.
In this work, we extend the Gaussian representation with reflective attributes and the learnable environmental map for reflective objects.

\textbf{Inverse Rendering.} 
It aims to decouple geometry, material, and lighting through a physically-based constraint model, given a set of images.
The pioneering work, Ref-NeRF~\cite{refnerf}, achieves realistic specular highlights and reflections, addressing the limitations of NeRF in rendering shiny objects. By leveraging a novel training approach that incorporates both geometry and appearance, Neus~\cite{wang2021neus} effectively captures fine details and complex shapes, enabling realistic surface reconstruction and rendering. 
By incorporating view-dependent effects directly into the neural implicit representation, Ref-Neus~\cite{refneus} achieves more accurate and visually appealing results for scenes featuring glossy and reflective materials.
NeRO~\cite{liu2023nero} utilizes a two-stage method to approximate the shading effects of both direct and indirect lights, achieving pleasing rendering results.
GaussianShader~\cite{jiang2024gaussianshader} proposes a shading function and a normal extraction method for 3D Gaussian Splatting. 
GS-IR~\cite{gsir} proposes a baking-based occlusion to construct indirect lighting and a depth derivation-based regularization for normal estimation.
3DGS-DR~\cite{3dgsdr} leverages a deferred rendering in pixel space to sample environmental effects and map them into the geometry surface.
However, previous methods reconstruct collapsed geometry that considerably limits the mapping of reflection and the learning of the HDR environment map. Our proposed RGS provides a simple and effective geometry-regularized solution to reconstruct more accurate geometry for specular mapping and achieve better inverse rendering.

%% file: sec/3_method.tex
\begin{figure*}[t!]
    \centering
    \includegraphics[width=0.98\linewidth]{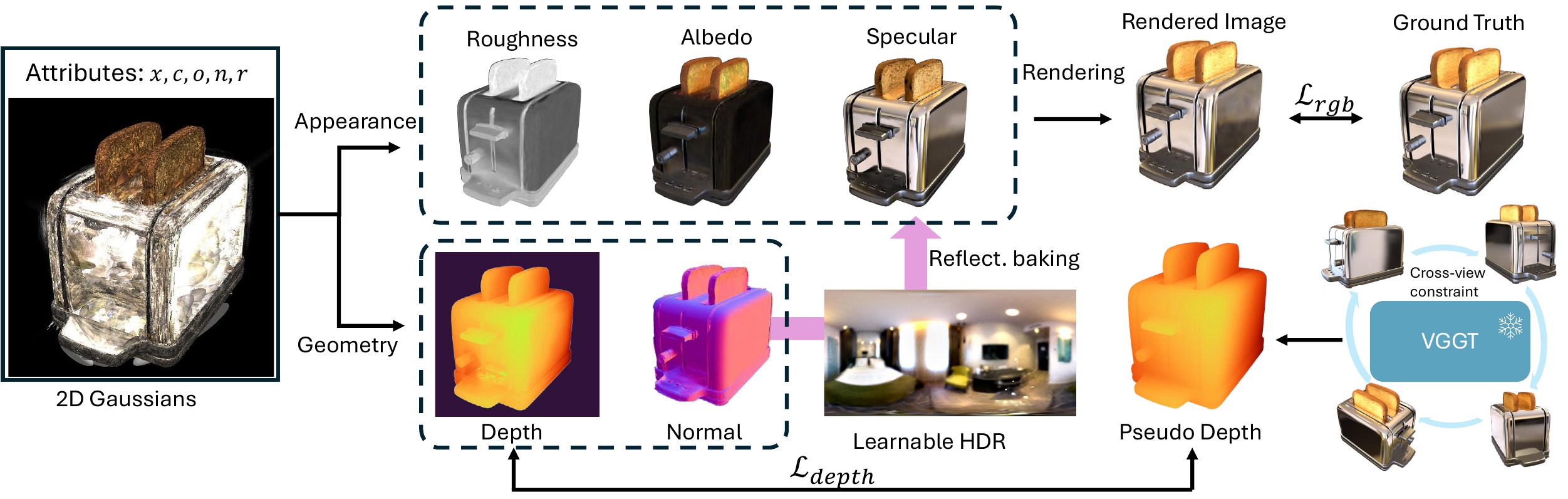} 
    \caption{\textbf{Pipeline of our proposed RGS.}
    Our method with the 2D Gaussian representation can render appearance and geometry maps with a learnable HDR map. 
    We utilize VGGT~\cite{wang2025vggt} inherent with cross-view constraints to provide accurate geometry regularization.
    RGS typically focuses on the specular capture for high-quality novel view synthesis. }
    \label{fig:method}
    \vspace{-5mm}
\end{figure*}

\section{Methodology}

We propose RGS, a novel deferred rendering-based reflection-aware Gaussian Splatting method that incorporates geometry regularization, specifically tailored for rendering reflective objects in novel views as shown in Fig~\ref{fig:method}. Our approach commences with the 2D Gaussian representation~\cite{2dgs} along the ray with direction $\mathbf{d_i}$ passing by $N_\text{d}$ 2D Gaussians. 
Computed by a weighted sum of these $N_\text{d}$ kernels, the per-pixel albedo color $\mathbf{C}(\mathbf{d}_i)$ can be obtained by integrating the colors $c_i(\mathbf{d}_i)$ with corresponding per-kernel spherical harmonics. Subsequently, we employ tile-based rasterization~\cite{kerbl2023gaussiansplatting} to blend the colors $c_i$ sorted by their respective depths $z_i$. 
The albedo color and depth can be represented as:
\begin{equation}
    \mathbf{C}(\mathbf{d}_i) = 
    \sum_{i = 1}^{N_\text{d}} w_i c_i(\mathbf{d}_i) ,~~\text{and}~~
     \mathbf{D}(\mathbf{d}_i) = 
    \sum_{i = 1}^{N_\text{d}} w_i  z_i(\mathbf{d}_i), 
\end{equation}
with the weight $w_i = \alpha_i \prod_{j=1}^{i-1}\left(1-\alpha_j\right)$, where the $\alpha_i$ is calculated with Gaussian opacity $o_i$ through:
\begin{equation}
    \alpha_i=o_i\exp(-\frac{1}{2}(x_i)^T \Sigma_i^{-1}(x_i)),
\end{equation}
where $x_i$ denotes the distance between the Gaussian center and the camera following direction $\mathbf{d}_i$. In addition to these basic rendering attributes, we also render the normal $\mathbf{N}(\mathbf{d}_i)$ and the roughness map $\mathbf{R}(\mathbf{d}_i)$. Each Gaussian is associated with attributes of a normal $n_i$ and roughness coefficient $r_i$, which contribute to $\mathbf{N}(\mathbf{d}_i)$ and $\mathbf{R}(\mathbf{d}_i)$ via
\begin{equation}
    \mathbf{N}(\mathbf{d}_i) = \sum_{i = 1}^{N_\text{d}} w_i (n_i + \Delta n_i) , \ 
    \mathbf{R}(\mathbf{d}_i) = \sum_{i = 1}^{N_\text{d}} w_i r_i(\mathbf{d}_i)  ~.
\end{equation}
 $n_i$ is the normal vector defined in 2DGS~\cite{2dgs}. $\Delta n_i$ serves as a regularization term to encourage $n_i$ closer to true normal.

\subsection{Physically-Based Shading and BRDF Modeling}
In this work, we utilize the deferred rendering framework to render RGB images following the physically-based shading and BRDF principles.
The final rendered RGB image can be formulated as:
\begin{align}
    \mathbf{C} &=  \mathbf{C}(\mathbf{d}_i) + 
 f_\phi(\mathbf{C}_s(\mathbf{d}_o)), \nonumber \\  
 \mathbf{d}_o &= 2 \frac{ \langle\mathbf{d}_i, \mathbf{N} (\mathbf{d}_i)\rangle}{\| \mathbf{N}(\mathbf{d}_i) \|} \mathbf{N}(\mathbf{d}_i) - \mathbf{d}_i,
\end{align}
where $f_\phi$ denotes a shader and $\mathbf{C}_s(\mathbf{d}_o)$ represents the learnable HDR, as outgoing radiance feature by reflection direction $\mathbf{d}_o$. $\mathbf{C}_s(\mathbf{d}_o)$ also means the specular feature that can be computed through:
\begin{equation}
\mathbf{C}_s(\mathbf{d}_o) = \int_{\Omega^+} f_r( \mathbf{d}_i, \mathbf{d}_o; \mathbf{R}) \cdot L_i(\mathbf{d}_i) \cdot (\mathbf{d}_i \cdot \mathbf{N}) \, d\mathbf{d}_i,
\end{equation}
which aims to compute the integral of reflection and incident light over the normal hemisphere. $f_r( \mathbf{d}_i, \mathbf{d}_o; \mathbf{R}) $ means the bidirectional reflectance distribution function (BRDF), which can be represented via:
\begin{equation}
\int_{\Omega^+} f_r( \mathbf{d}_i, \mathbf{d}_o; \mathbf{R}) = \frac{D(\mathbf{h}; \mathbf{R}) \cdot F(\mathbf{d}_i, \mathbf{h}) \cdot G(\mathbf{d}_i, \mathbf{d}_o, \mathbf{h})}{4 (\mathbf{n} \cdot \mathbf{d}_i)(\mathbf{n} \cdot \mathbf{d}_o)},
\end{equation}
where $\mathbf{h} = \frac{\mathbf{d}_i + \mathbf{d}_o}{\|\mathbf{d}_i + \mathbf{d}_o\|} $ means the half vector. $D$ denotes the microfacet distribution function, controlled by the roughness.
$F$ and $G$ represent the Fresnel term, and the shadowing-masking term, respectively.
Therefore, the rendering color loss can be formulated as:
\begin{equation}
    \mathcal{L}_\text{rgb} = 
    \lambda \mathcal{L}_1(\mathbf{C}, \mathbf{C}_\text{gt}) +
    (1-\lambda)\mathcal{L}_\text{DSSIM}(\mathbf{C}, \mathbf{C}_\text{gt}),
\label{eq:loss_photometric}
\end{equation}
with a pre-defined weight $\lambda$ balancing $\mathcal{L}_1$ and $\mathcal{L}_\text{DSSIM}$ as mentioned in 3DGS~\cite{kerbl2023gaussiansplatting}. 
In addition, we also render $\Delta n_i$ to obtain its 2D map for later constraint
$
    \Delta\mathbf{N}(\mathbf{d}_i) = 
    \sum_{i=1}^{N_\text{d}} w_i  \Delta n_i, 
$
where $\Delta\mathbf{N}(\mathbf{d}_i)$ is the 2D map of $\Delta n_i$. That can be seen as a residual form of the original normal map. To promote the accuracy of 2D Gaussian normal vectors, we regularize $\Delta n_i$ through:
\begin{equation}
     \mathcal{L}_{\Delta n}  =  \| \Delta\mathbf{N}(\mathbf{d}_i) \|_2,
\end{equation}
where it is designed to encourage 2D Gaussian normal for more precise convergence, thereby reducing reliance on the residual normal for guidance.



\subsection{Cross-view Shape Consistency Regularization}
\label{sec:method:sdr}
To reconstruct fine-grain geometry, we introduce a powerful foundation model, VGGT~\cite{wang2025vggt}, that possesses a strong geometry prior and cross-view constraint for the estimation of depth.
As shown in Fig.~\ref{fig:method}, we leverage VGGT to estimate the depth given all training views. 
In this way, we can obtain a pseudo depth map $\mathbf{D}'$ since VGGT is not the metric scale method.
Even though these depth images do not possess physical meaning, they can provide cross-view geometry consistency constraints to regularize Gaussian Splatting.
Due to the scale ambiguity, we use the Pearson loss~\cite{benesty2009pearson}, a correlation-based criterion, to regularize the depth map rendered by 2D Gaussians: 
\begin{equation}
    \mathcal{L}_\textnormal{depth}(\mathbf{D}', \mathbf{D} )= 1 -\frac{\text{Cov}(\mathbf{D}',  \mathbf{D} )}{\sigma(\mathbf{D}') \sigma(\mathbf{D} )},
\end{equation}
where Cov($\cdot$) calculates covariance of two variables and $\sigma$ means a variance operator. $\mathbf{D} $ means the depth map rendered by Gaussian Splatting.

In our proposed cross-view depth regularization strategy, the obtained pseudo depth map is estimated from the powerful 3D foundation model, VGGT.
Note that this strategy benefits from the multiple views to impose cross-view consistency that can refine geometry at a fine-grain level. 
In particular, we employ a Pearson loss to regularize the depth map rendered by 2D Gaussians since VGGT cannot estimate real depth. The pseudo depth maps essentially contain accurate depth information, providing effective cross-view consistency information for the regularization of 2D Gaussians. In this way, our RGS can reconstruct more accurate geometry for better specular effect mapping.

\noindent\textbf{Reflection-guided Normal Regularization.}
In our experiments, we find that reconstructed reflective regions always collapse since Gaussian-based methods are prone to overfitting specular effects and reconstructing cavernous geometry on reflective regions. 
Therefore, it is necessary to regularize surface geometry for better mapping the specular effects. 
Inspired by Total Variation (TV) loss~\cite{tvloss}, we find that it has a powerful regularization capacity for surface geometry. 
Consequently, we employ the rendered roughness map $\mathbf{R}$ to pinpoint reflective regions $ \hat{\beta}$ through:
\begin{align}
    \hat{\beta} =
    \begin{cases}
    1, & \text{if } \mathbf{R} \leq \mathbf{R}_p \\
    0, & \text{otherwise}
    \end{cases},
\end{align}
where $\mathbf{R}_p$ means the p-\textit{th }percentile of $\mathbf{R}$ and $\hat{\beta}$ can be seen as a mask, identifying strongly reflective regions. Empirically, $\mathbf{R}_p$ is enough to capture reflective regions.  Therefore, we leverage TV loss to regularize the normal maps $\mathbf{N}$ within strongly reflective regions via:
\begin{equation}
    \mathcal{L}_\text{normal} =  \hat{\beta} \odot \sum_{i,j} \| \mathbf{N}_{i+1,j} - \mathbf{N}_{i,j}\|^2 + \|\mathbf{N}_{i,j+1} - \mathbf{N}_{i,j}\|^2, 
\end{equation}
where $\odot$ denotes Hadamard product. In this way, the surface collapse problem in reflective regions would be largely mitigated, leading to better specular mapping.


\begin{figure}[t!]
  \centering
    \includegraphics[width=\linewidth]{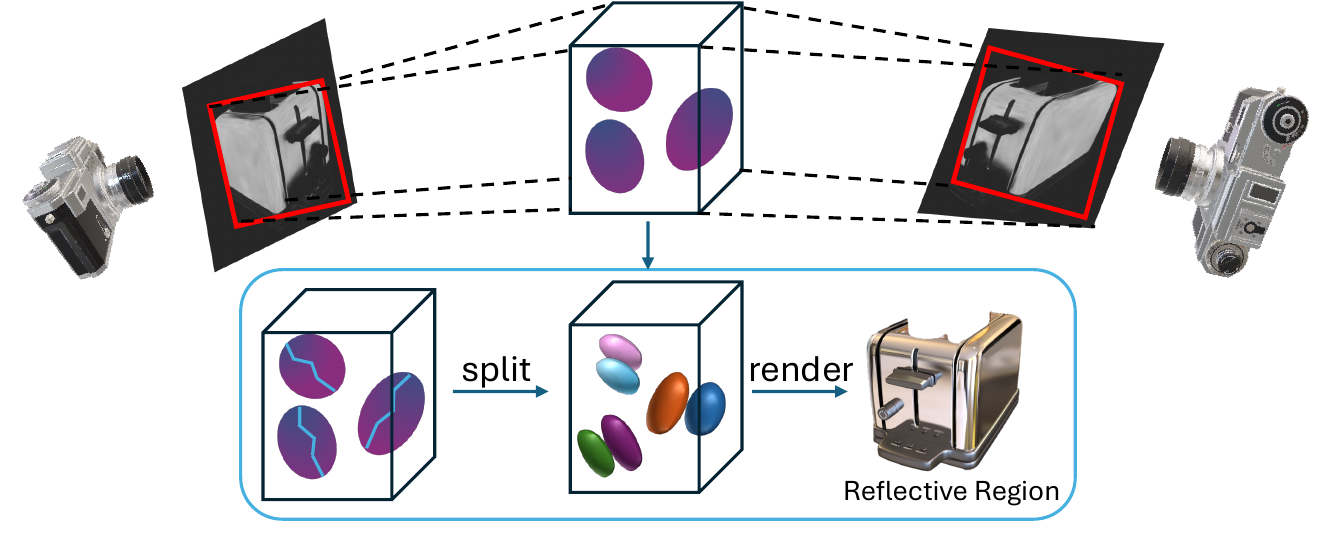}
    \captionof{figure}{\textbf{Illustration of the proposed reflection-guided densification.} We utilize a multi-view ray tracing technique to cast rays from the vertices of the red rectangle for localizing significant 2D Gaussians and enhancing their densification for better specular pattern capture.}
    \label{fig:method_densify}
  \vspace{-5mm}
\end{figure}

\begin{algorithm}[t]

\caption{Reflection-guided Densification }
\label{algorithm:rd}

\KwIn{Multi-view camera parameters $\{\mathcal{C}_v\}_{v=1}^V$, current 3D Gaussians $\mathcal{G}$}
\KwOut{Updated 3D Gaussians $\mathcal{G}'$ with enhanced specular fitting}
\SetKwFor{ForEach}{for each}{do}{end}
\ForEach{view $v \in \{1, \dots, V\}$}{
    Extract specular intensity map $\hat\beta$;
    Compute 2D bounding boxes $\mathcal{B}_v$ on $\hat\beta$;
    
    Emit rays from vertices of $\mathcal{B}_v$ using camera parameters $\mathcal{C}_v$\;
}

\ForEach{pair of views $(v_i, v_j)$}{
    Intersected planes formed by rays from $\mathcal{B}_{v_i}$ and $\mathcal{B}_{v_j}$ to localize 3D regions $\mathcal{R}$; 
    
    Identify Gaussians $\mathcal{G}_{\text{spec}} \subset \mathcal{G}$ located in regions $\mathcal{R}$\;
}

\ForEach{Gaussian $g \in \mathcal{G}_{\text{spec}}$}{
    Decrease densification threshold $\tau_g \leftarrow \tau_g / 2$\;
}

Update $\mathcal{G}'$ by performing densification using updated thresholds\;

\Return{$\mathcal{G}'$}

\end{algorithm}

\subsection{Reflection-guided Densification}
\label{sec:method:rd}

Although some previous methods can reconstruct plausible geometry, they fail to capture strong specular.
Specular, as a kind of view-dependent information, is challenging for a limited number of explicit 2D Gaussians to capture its patterns.
Inspired by MVGS~\cite{du2024mvgs}, which leverages a multi-view densification technique to improve the densification effect, we propose a reflection-guided densification strategy to foster 2D Gaussians learning toward reflective 3D regions and facilitate the fitting convergence for solving this problem. 

As shown in Fig.~\ref{fig:method_densify}, we start by locating the 2D Gaussians that contribute significantly to this kind of specular information, and densify these 2D Gaussians. 
To achieve this, we extract specular intensity maps from roughness maps, which highlight 2D reflective regions. Then we employ a cross-view ray tracing technique, emitting rays from 2D reflective regions and locating significant Gaussians in 3D space. Finally, we enhance the densification effect of these 2D Gaussians.

To be specific, we first calculate the bounding boxes of 2D reflective maps, representing the highly reflective regions of an object, as shown in the red rectangle of Fig.~\ref{fig:method_densify}. 
Instead of emitting rays from these reflective regions, we leverage the camera parameters to emit rays in the vertices of bounding boxes per view, avoiding huge computational costs.
These rays forming planes would have intersections with others emitted from other views. The intersected planes can locate 3D areas containing a set of important 2D Gaussians contributing significantly to the specular. We enhance these 2D Gaussians' densification by decreasing their densification threshold by half. In this way, the 2D Gaussians in these regions are easier to learn and capture the specular. It leads to a better fitting process and a more compact structure for 2D Gaussians. Therefore, our RGS can be more effective in capturing specular effects and rendering specular highlights. The proposed reflection-guided densification process is described in the Algorithm~\ref{algorithm:rd}.

\subsection{Losses and Implementation}
Our method is designed in a simple and streamlined manner. In summary, the total loss of RGS can be represented as:
\begin{equation}
    \mathcal{L} =\lambda_\text{rgb} \mathcal{L}_\text{rgb} + \lambda_\text{depth} \mathcal{L}_\text{depth} +  
    \lambda_\text{normal} \mathcal{L}_\text{normal} +
      \lambda_{\Delta n} \mathcal{L}_{\Delta n},
\end{equation}
where $\mathcal{L}_\text{rgb}$ is our main loss function that contributes considerably to the whole learning process. $\mathcal{L}_\text{normal}$ and $\mathcal{L}_\text{depth}$ are the regularization terms that regularize the surface to be smooth, avoiding overfitting phenomena and collapsed hollows. $\mathcal{L}_{\Delta n}$ is a regularization function that facilitates more accurate learning of the normal vectors from 2D Gaussians.

\noindent\textbf{Implementation.}
First, we set the weight $\lambda_\text{rgb}$ as 1 for the main RGB loss in our experiments. Then, the weights for geometry-regularized losses are set as $\lambda_\text{normal}=1\text{e}^{-2}$ and $\lambda_\text{depth} = 1\text{e}^{-3}$. 
For the residual normal loss weight, we set $\lambda_{\Delta n}=1\text{e}^{-3}$. To obtain the reflection intensity map, we set the percentile $p$ as 0.9. As for other parameter settings, we follow 3DGS-DR~\cite{3dgsdr} for fair comparisons. We also leverage multi-view training~\cite{du2024mvgs} to enhance multi-view constraints and further improve performance. We conduct experiments on an RTX 3090 Ti GPU.

%% file: sec/4_exp.tex
\input{table/t1}

\input{table/t2}

\input{table/t_normal}

\begin{figure}[t!]
    \centering
    \includegraphics[width=\linewidth]{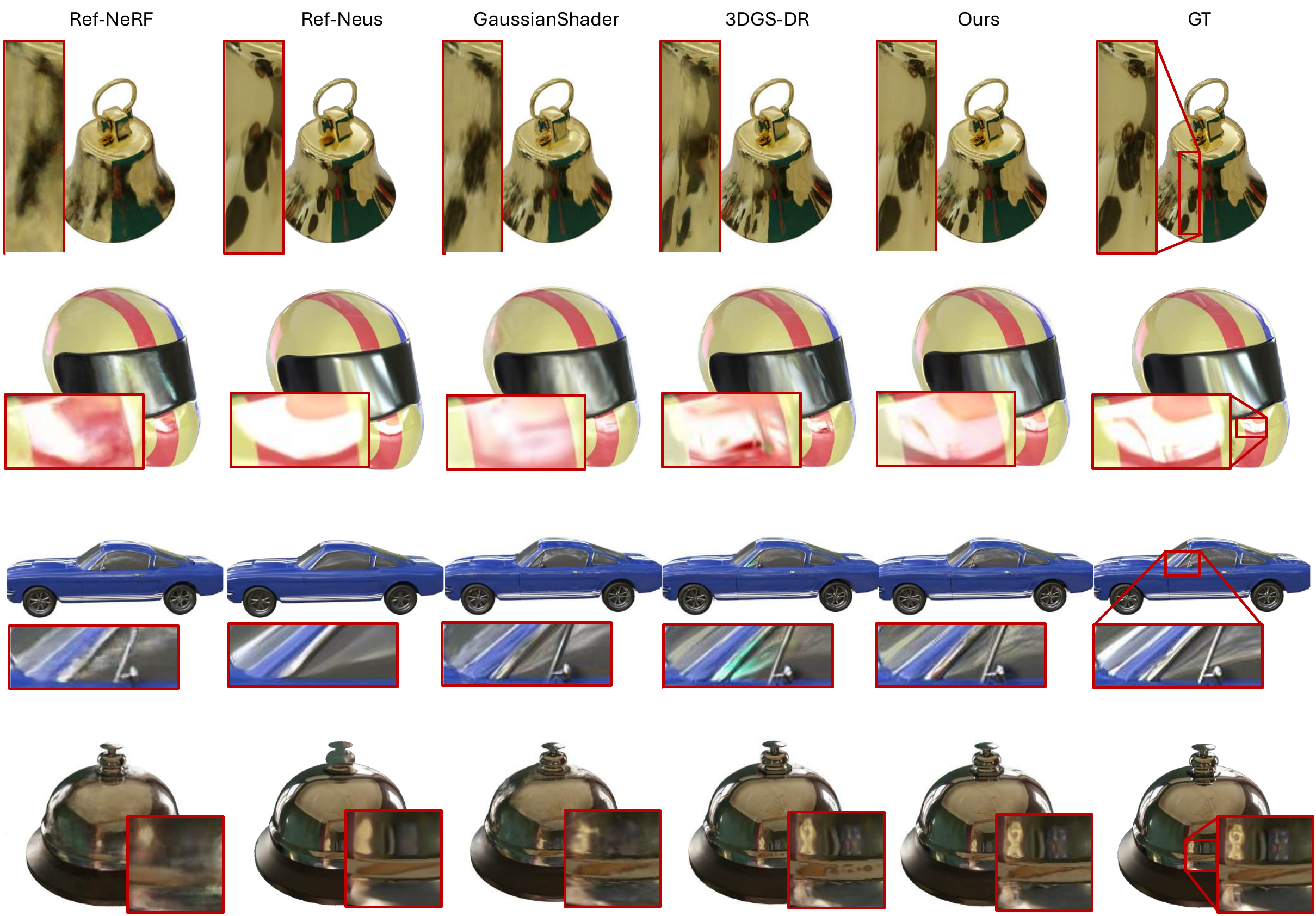} 
    \caption{\textbf{Visual comparisons of our method with state-of-the-art methods for novel view synthesis toward challenging reflective objects. } We demonstrate that our method reconstructs specular and reflective surfaces more precisely.
    We display the close-up for better differentiation.
 }
    \label{fig:refl}
    \vspace{-5mm}
\end{figure}

\begin{figure}[t!]
    \centering
    \includegraphics[width=0.99\linewidth]{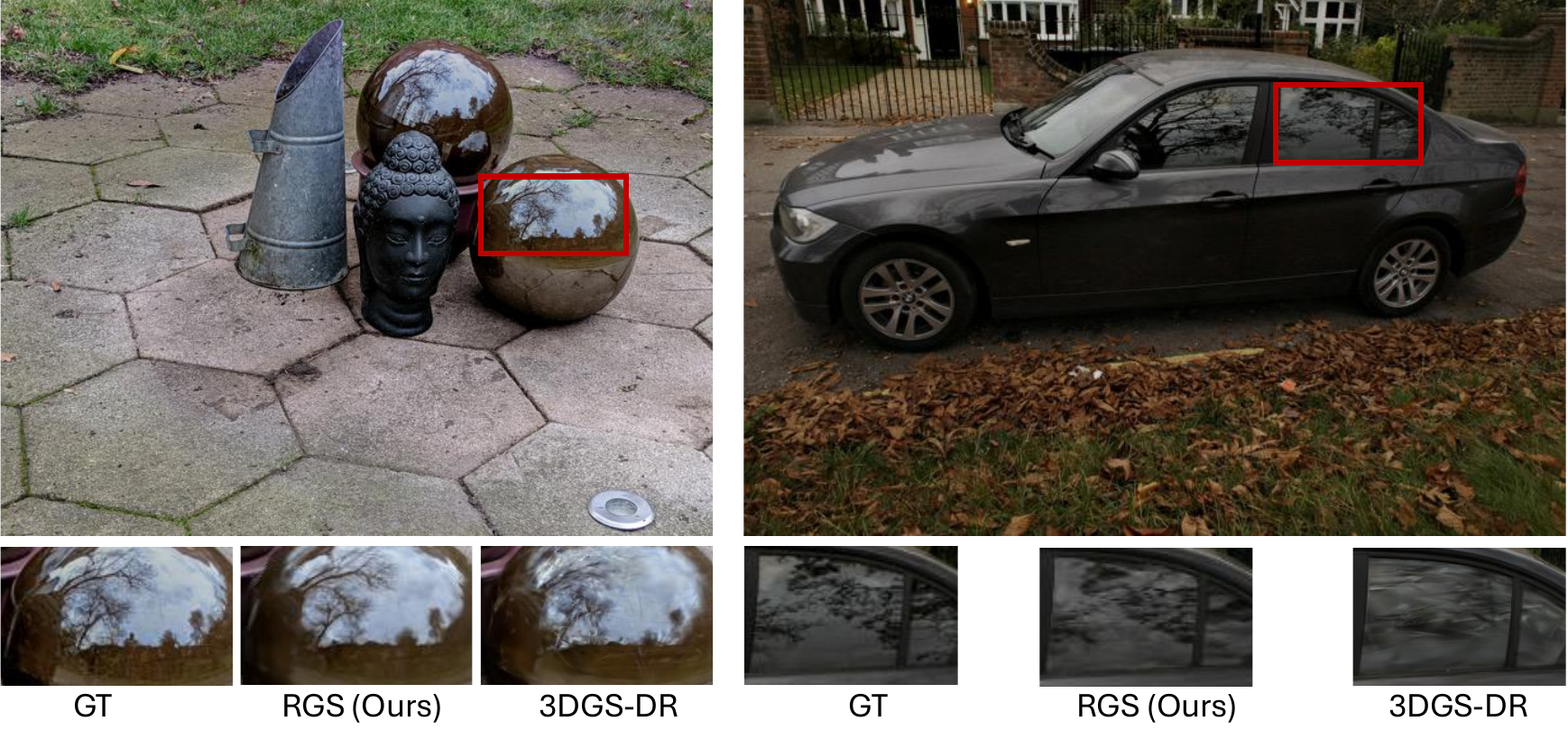} 
    \caption{\textbf{Qualitative comparisons of our method with existing state-of-the-art methods on Ref-NeRF Real~\cite{refnerf}. } We demonstrate that our method not only improves novel view synthesis results of synthetic reflective objects but also real-world reflective objects.
 }
    \label{fig:real}
    \vspace{-5mm}
\end{figure}

\begin{figure}[t!]
    \includegraphics[width=0.95\linewidth]{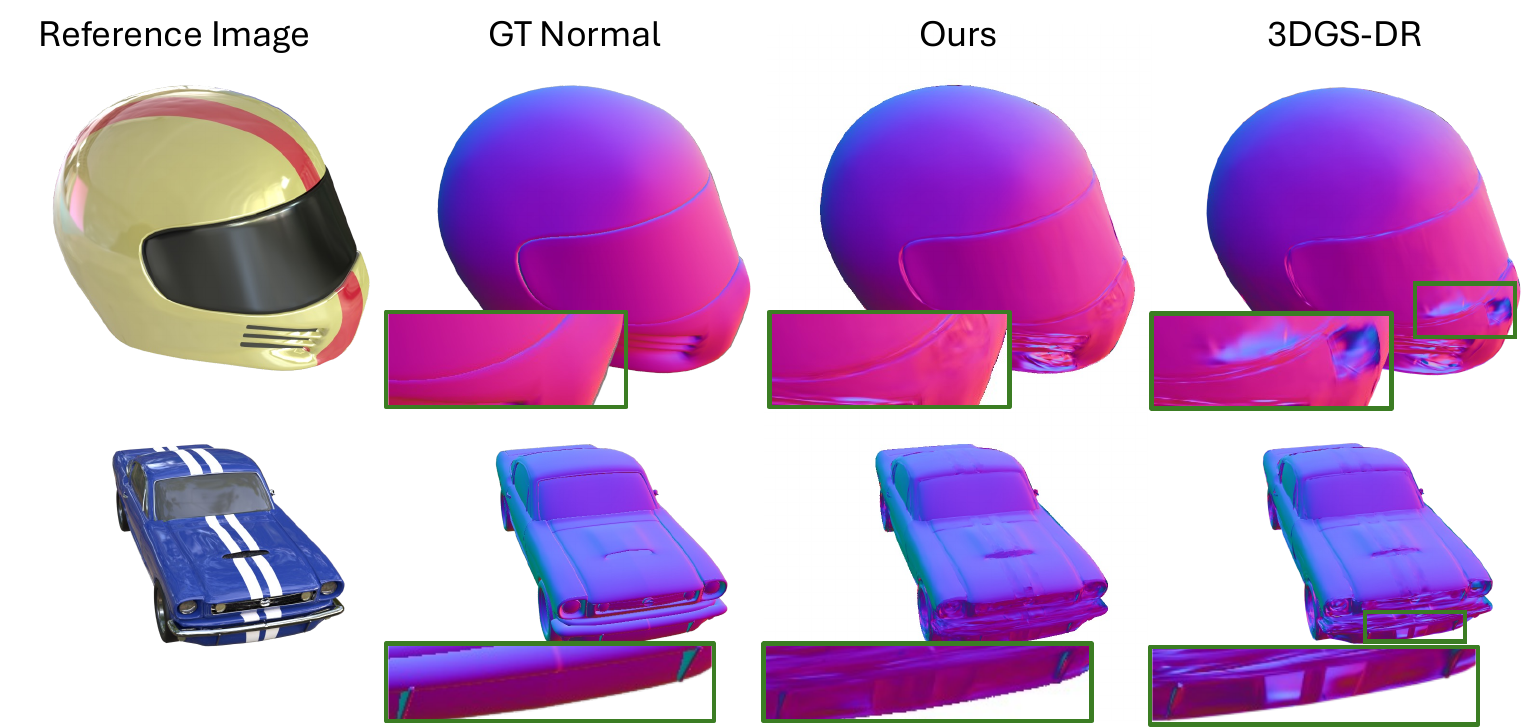} 
    \caption{\textbf{Comparisons of rendered normal between our proposed method and 3DGS-DR. } Better normal means more accurate geometry reconstruction.
 }
    \label{fig:normal}
\vspace{-3mm}
\end{figure}

\section{Experiments}

\noindent\textbf{Datasets and Metrics.} To demonstrate the effectiveness of our method to reflective objects and applicability to general objects, we conduct extensive experiments on four datasets with 23 scenes, including reflective and general object datasets, such as Shiny Blender~\cite{refnerf}, Glossy Synthetic~\cite{liu2023nero}, Ref-NeRF Real~\cite{refnerf}, and NeRF Synthetic~\cite{mildenhall2021nerf}. For the evaluation metrics, we follow previous works, utilizing PSNR, SSIM, and LPIPS to evaluate the performance of novel view synthesis.

\subsection{Quantitative and Qualitative Results}
In our experiments, we evaluate our method for novel view synthesis quantitatively and qualitatively, especially for challenging reflective objects.

\noindent\textbf{Novel View Synthesis.}
Novel view synthesis measures the performance of a reconstructed 3D model synthesizing various views distinct from training views.
In our experiments, we conduct extensive comparisons from reflective objects to general objects, demonstrating the generalization of our method. As shown in Table~\ref{t1_refl}, we compare our method with existing state-of-the-art novel view synthesis methods for reflective objects, including Ref-NeRF~\cite{refnerf}, NPC~\cite{npc}, 3DGS~\cite{kerbl2023gaussiansplatting}, GaussianShader~\cite{jiang2024gaussianshader},
ENVIDR~\cite{liang2023envidr}, 3DGS-DR~\cite{3dgsdr}, Spec-Gaussian~\cite{yang2024spec}, 3iGS~\cite{tang20243igs},  EnvGS~\cite{xie2024envgs}, Ref-GS~\cite{zhang2024ref}, and Ref-Gaussian~\cite{yao2024reflective} on three representative datasets, such as Shiny Blender~\cite{refnerf}, Glossy Synthetic~\cite{liu2023nero}, and Ref-NeRF Real~\cite{refnerf}. 
Note that these datasets contain reflective objects with glossy surfaces. Ref-NeRF Real~\cite{refnerf} is a real dataset containing reflective objects in real scenes, so the real lighting condition would be more challenging.
As we can see from Table~\ref{t1_refl}, our proposed RGS achieves the best quantitative results compared with other methods, demonstrating that our method achieves state-of-the-art performance.
We also display visual comparisons in Fig.~\ref{fig:refl} and Fig.~\ref{fig:real}.
These visual comparisons demonstrate that our proposed RGS can reconstruct better specular and clearer reflective appearance close to the ground truth.
To general objects, we present our results in Table~\ref{t2_nerf}. Table~\ref{t2_nerf} demonstrates our method achieving state-of-the-art results compared with other methods. It also indicates our method not only performs well for reflective objects but also for general objects.
It is attributed to the proposed RGS focusing on geometry regularization and 2D Gaussian densification for specular capture.

\noindent\textbf{Geometry Measurement.}
In addition to the RGB renderings, we also show geometry, as the normal map in Fig.~\ref{fig:normal} and Table~\ref{t_normal}.
As we can see in Fig.~\ref{fig:normal}, 3DGS-DR~\cite{3dgsdr} fails to reconstruct the accurate normal. The inaccurate normal leads to the project error of specular and affects the learning of the environment light.
In contrast, our proposed RGS achieves high-quality normal reconstruction.
We also show the quantitative results in Table~\ref{t_normal}. It objectively demonstrates that our proposed RGS can achieve better normal reconstruction compared with previous methods, in terms of MAE, SSIM, and LPIPS.
The accurate normal suggests better geometry. It also indicates that our method can reconstruct more accurate surfaces to achieve better specular mapping for reflective object novel view synthesis.

\input{table/t_time}

\noindent\textbf{Budget Comparisons.}
We also conduct budget comparisons as shown in Table~\ref{t_budget}. Compared to 3DGS, our method achieves a better trade-off between rendering speed and performance. Note that our method can still reach the real-time rendering speed.  Even though our method utilizes the enhanced densification technique, the multi-view constraints encourage the structure of 3D Gaussians to be more compact. Therefore, our method can achieve state-of-the-art results with fewer parameters compared with previous methods.

\subsection{Ablation Study}
\begin{figure}[t!]
    \centering
    \includegraphics[width=0.99\linewidth]{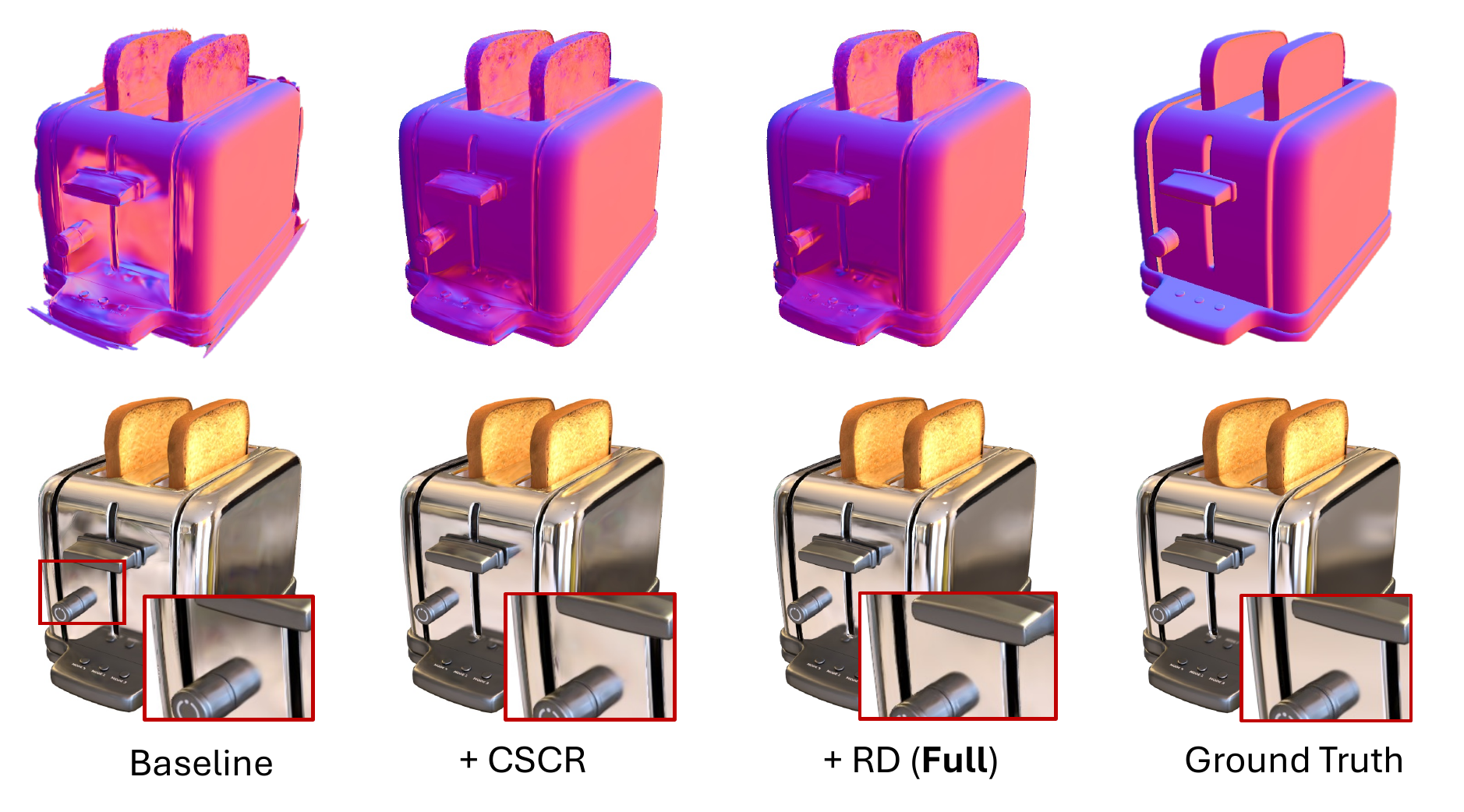} 
    \caption{\textbf{Ablation studies of our proposed components in RGS.} We show qualitative and quantitative results to fully demonstrate the effectiveness of RGS.
    CSCR represents the proposed cross-view shape consistency regularization.
        RD denotes the proposed reflection-guided densification.
    We use \textcolor{red!80}{red} rectangles to highlight the details for better differentiation.
 }
    \label{fig:ablation}
    \vspace{-5mm}
\end{figure}

To demonstrate the effectiveness of the proposed components, we conduct ablation studies on the Shiny Blender Dataset~\cite{refnerf}. 
In particular, we leverage 2DGS~\cite{3dgsdr} integrated with the deferred rendering framework as our baseline and progressively add our proposed components into it.
The ablation results are displayed in Fig~\ref{fig:ablation} and Table~\ref{t_ablation}.
It is obvious that the baseline produces a low-quality normal map with a collapsed surface geometry and also renders an inferior novel view. 
On the contrary, when we integrate our proposed cross-view shape consistency regularization (CSCR) into it, the normal map becomes smooth without collapse, and the novel view is approaching the Ground Truth. It is attributed to the proposed CSCR exploiting the foundation model prior and cross-view constraints to regularize the fitting process of 2DGS.
Finally, when we embed our reflection-guided densification (RD) into the method, our method is completed and achieves high-quality results compared with previous methods. 
It makes the specular learning process easier with the ray-based densification strategy.
In particular, the normal map is very similar to Ground Truth, which indicates that our method achieves highly accurate geometry.
These ablation studies demonstrate the effectiveness of the proposed components.

\input{table/t_ablation}

\begin{figure}[t!]
  \centering

     \includegraphics[width=0.95\linewidth]{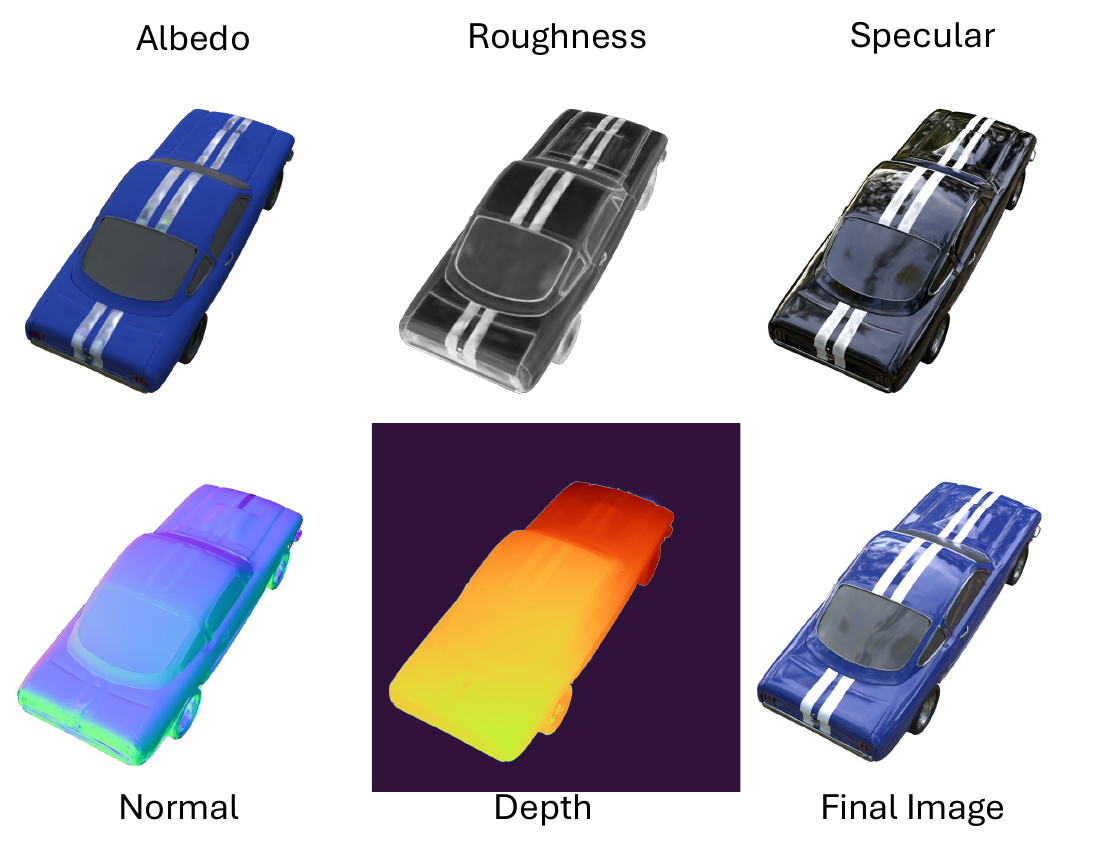} 
    \captionof{figure}{\textbf{Decomposition results of our proposed RGS.} The final image is the combination of albedo and specular maps.}
    \label{fig:decompose}
     \vspace{-4mm}
\end{figure}

\noindent\textbf{Decomposition Results.} To further analyze our method, we visualize our rendering attributes, including albedo, roughness, specular, depth, normal, and the final image as shown in Fig.~\ref{fig:decompose}.
 As we can see from the top of Fig.~\ref{fig:decompose}, our RGS can successfully decouple albedo and specular, which facilitates for the learning and mapping of the environmental lighting effect.  In addition, the bottom of Fig.~\ref{fig:decompose} demonstrates that rendered depth and normal maps are accurate in representing the learned geometry under our proposed restriction.

\begin{figure}[t!]
    \centering
    \includegraphics[width=0.95\linewidth]{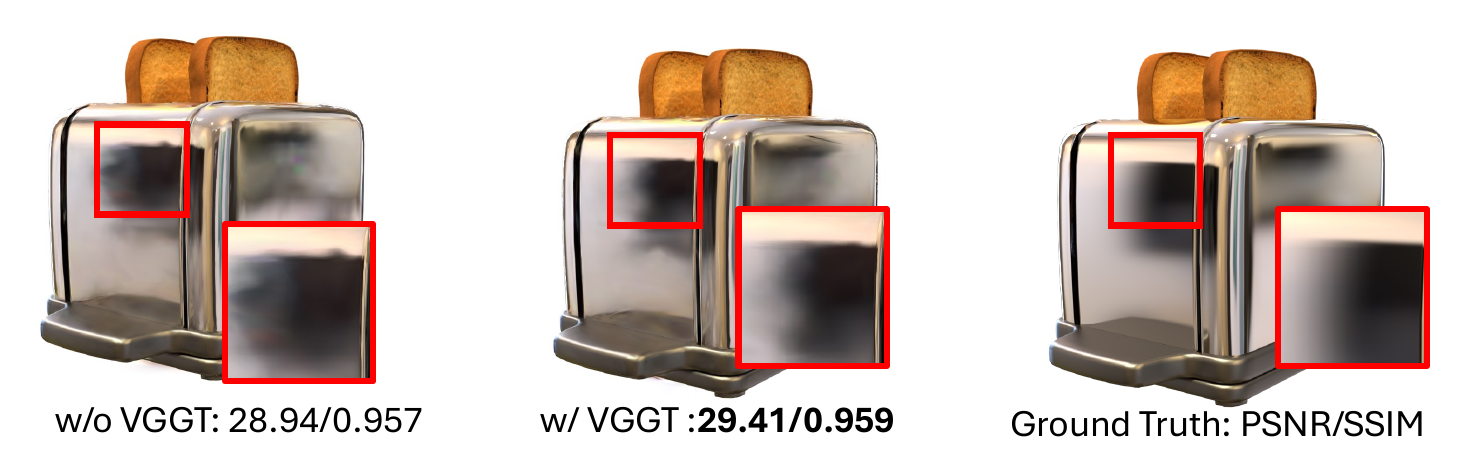} 
    \captionof{figure}{\textbf{Ablation study of foundation model assistance.} We demonstrate that VGGT can provide helpful geometry supervision for Gaussian Splatting to improve performance. }
    \label{fig:ablation_pose}

  \vspace{-5mm}
\end{figure}

\noindent\textbf{Foundation Model Assistance.}
We also conduct an ablation study in Fig.~\ref{fig:ablation_pose} to demonstrate the necessity of the involvement of the foundation model, VGGT.  As displayed in Fig.~\ref{fig:ablation_pose}, when the depth assistance supervision is involved, our RGS can reconstruct more accurate geometry for specular mapping and improve novel view synthesis performance. 
It is attributed to the accurately estimated depth from VGGT, which contains powerful cross-view constraints and the foundation model prior.
These results highlight the potential of foundation models to improve geometry reconstruction, providing stronger constraints for training 3D Gaussians and enabling high-quality novel view synthesis.

%% file: table/t1.tex
\begin{table*}[t!]
\caption{\textbf{Quantitative comparisons of state-of-the-art reflective object novel view synthesis methods.} We demonstrate our method effectively improves novel view synthesis performance for challenging reflective objects even in real scenes.  We display results on Glossy Synthetic datasets~\cite{liu2023nero}, Shiny Blender~\cite{refnerf}, and Ref-NeRF Real~\cite{refnerf}.  The \colorbox{red!30}{best}, \colorbox{orange!30}{second best}, and \colorbox{yellow!30}{third best} results are highlighted respectively for better visibility. }
\centering

\resizebox{0.99\linewidth}{!}
{
    \begin{tabular}{ l|ccc|ccc|ccc}
    \toprule
    Dataset & \multicolumn{3}{c|}{Glossy Synthetic} & \multicolumn{3}{c|}{Shiny Blender} & \multicolumn{3}{c}{Ref-NeRF-Real} \\
    \midrule \midrule 
    Method \& Metrics& PSNR$\uparrow$ & SSIM$\uparrow$ & LPIPS$\downarrow$  & PSNR $\uparrow$ & SSIM$\uparrow$ & LPIPS$\downarrow$ & PSNR$\uparrow$ & SSIM$\uparrow$ &LPIPS$\downarrow$   \\
      \midrule 
    Ref-NeRF~\cite{refnerf} & 27.49& 0.927& 0.100 & 33.12& 0.961& 0.079& 23.62& 0.645&\cellcolor{orange!30}0.238 \\
    NPC~\cite{npc}& 21.96& 0.841& 0.181 & 27.48& 0.921& 0.145& 22.87& 0.605&0.320 \\

    3DGS~\cite{kerbl2023gaussiansplatting}& 26.49&0.917&0.092 &30.35&0.946&0.083& 23.85&  \cellcolor{yellow!30}0.659& \cellcolor{red!30}0.230 \\

   3iGS~\cite{tang20243igs}& 26.39&0.913&0.089 &30.64&0.948&0.077& 23.95&  0.657& 0.258 \\

    GaussianShader~\cite{jiang2024gaussianshader}& 27.53&0.921&0.086 & 31.96&0.957&0.067& 23.46& 0.647&\cellcolor{yellow!30}0.257 \\

    ENVIDR~\cite{liang2023envidr}& 29.56&  0.952& 0.059 & 
 33.46&0.967  &\cellcolor{red!30}0.045& 23.00& 0.605&0.331 \\

    Spec-Gaussian~\cite{yang2024spec} & 27.54&  0.931& 0.079 & 
 31.00 & 0.950  & 0.075& 23.71& 0.652&0.259 \\

    3DGS-DR~\cite{3dgsdr} &30.13&0.953& \cellcolor{yellow!30}0.058 & 34.08&0.971&\cellcolor{yellow!30}0.052&  23.99 &0.658 & 0.264     \\
    EnvGS~\cite{xie2024envgs} &  29.78 &   0.952 &   0.061& 33.82 &   0.968  & 0.065  &  \cellcolor{yellow!30}24.15 & \cellcolor{yellow!30}0.659 & 0.259    \\

    Ref-GS~\cite{zhang2024ref}   &  \cellcolor{orange!30}30.59 &   \cellcolor{yellow!30}0.957 &   \cellcolor{yellow!30}0.058 & \cellcolor{yellow!30}34.80 &   \cellcolor{orange!30}0.973  & 0.056  &  \cellcolor{orange!30}24.21 & \cellcolor{orange!30}0.661 & 0.258    \\
    
    Ref-Gaussian~\cite{yao2024reflective} & \cellcolor{yellow!30}30.36 & \cellcolor{orange!30}0.958 & \cellcolor{orange!30}0.051 &    \cellcolor{orange!30}35.01  &   \cellcolor{yellow!30}0.972& 0.056&  24.13 & 0.657 & 0.258    \\

    \midrule
    RGS (\textbf{Ours}) & \cellcolor{red!30}31.24 & \cellcolor{red!30} 0.965 & \cellcolor{red!30} 0.043 &  \cellcolor{red!30}35.38  & \cellcolor{red!30}0.976& \cellcolor{orange!30}0.050& \cellcolor{red!30}24.31& \cellcolor{red!30}0.670 & \cellcolor{yellow!30}0.257

     \\
    
    \bottomrule
    \end{tabular}
}

\label{t1_refl}
\vspace{-5mm}
\end{table*}

%% file: table/t2.tex
\begin{table}[t]
\centering
\caption{\textbf{Quantitative evaluation of our method compared with previous methods on the NeRF Synthetic dataset}~\cite{mildenhall2021nerf}.
We demonstrate our method not only performs well for reflective objects but also has a good generalization to general objects. 
GShader* means that we report the results with the official implementation repository of GaussianShader.
}
\resizebox{0.99\linewidth}{!}
{
\begin{tabular}{l|ccccccc}
\toprule
Method & NeRF\cite{mildenhall2021nerf} & VolSDF\cite{volsdf} & Ref-NeRF\cite{refnerf} & ENVIDR\cite{liang2023envidr} & 3DGS\cite{kerbl2023gaussiansplatting} & GShader*\cite{jiang2024gaussianshader} & RGS (\textbf{Ours})        \\ \midrule \midrule
PSNR   & 31.01                                                & 27.96                                      & 31.29                                         & 28.13                                               & \cellcolor{orange!30}33.30                  & \cellcolor{yellow!30}31.46                          & \cellcolor{red!30}33.51    \\ 
SSIM   & 0.947                                                & 0.932                                      & 0.947                                         & 0.956                                               & \cellcolor{red!30}0.969                     & \cellcolor{yellow!30}0.960                          & \cellcolor{orange!30}0.968 \\ 
LPIPS  & 0.081                                                & 0.096                                      & 0.058                                         & 0.067                                               & \cellcolor{orange!30}0.030                  & \cellcolor{yellow!30}0.042                          & \cellcolor{red!30}0.028    \\ \bottomrule
\end{tabular}
}
\label{t2_nerf}

\vspace{-2mm}
\end{table}

%% file: table/t_normal.tex
\begin{table}[t!]

\caption{\textbf{Quantitative normal comparisons among different methods on the Shiney Blender dataset~\cite{refnerf}.} We demonstrate that our method can reconstruct better normal maps for more precise geometry.}
\label{t3}
\centering
\scriptsize
\def\arraystretch{1.2}

\setlength{\tabcolsep}{0.8em}{
\begin{tabular}{lccccc}
\toprule
Method            & ENVIDR  &  Ref-Gaussian  & 3DGS-DR &EnvGS & RGS (Ours) \\ \midrule \midrule
MAE  $\downarrow$       &  2.74 & 2.18  &  2.62    &  2.66  &   \textbf{2.04 }\\
SSIM $\uparrow$  & 0.948 & 0.952 &   0.947 &  0.945   &\textbf{0.954}  \\
LPIPS $\downarrow$        &  0.072 &  0.053   &   0.065     &  0.062    &     \textbf{0.051}   \\ \bottomrule
\end{tabular}
\label{t_normal}
}

\vspace{-5mm}

\end{table}

%% file: table/t_time.tex
\begin{table}[t!]
  
\caption{\textbf{Budget comparisons among different methods on Shiney Blender dataset.} We demonstrate that our method with fewer parameters achieves state-of-the-art performance.}
\label{t3}
\centering

\scriptsize
\def\arraystretch{1.2}
\setlength{\tabcolsep}{0.8em}{
\begin{tabular}{lcccc}
\toprule
Method            & 3DGS  & 3DGS-DR &EnvGS & RGS (Ours) \\ \midrule \midrule
Parameters $\downarrow$       & 3.3M & 1.1M   & 2.9M & \textbf{0.1M }    \\
Train. Time  $\downarrow$       &  \textbf{0.6h }     &  2h     &  3h  &  3h  \\
FPS $\uparrow$  & \textbf{604} &   71 &    21  &   146 \\
PSNR $\uparrow$        &    30.35   &   34.08     & 33.82 &  \textbf{35.38 }          \\ \bottomrule
\end{tabular}
\label{t_budget}
}

\vspace{-5mm}

\end{table}

%% file: table/t_ablation.tex



\begin{table}[t!]

\caption{\textbf{Ablation study of our proposed RGS on the Shiney Blender dataset~\cite{refnerf}.}  We demonstrate the effectiveness of the proposed each component.}
\label{t3}
\centering
\scriptsize
\def\arraystretch{1.2}

\resizebox{0.99\linewidth}{!}
{
\begin{tabular}{cccccc}
\toprule
CSCR &  RD & PSNR& SSIM & FPS  & Parameters  \\ \midrule \midrule
  &     & 33.29& 0.967 &132 &1.4M  \\
\checkmark    &     & 34.54& 0.974  & 137  &0.4M    \\
   & \checkmark    & 34.63& 0.975  & 143 & 0.2M  \\
\checkmark   & \checkmark    & \textbf{35.38}& \textbf{0.976}  & \textbf{146} & 0.1M \\ \bottomrule
\end{tabular}
\label{t_ablation}
}

\vspace{-4mm}

\end{table}

%% file: sec/5_conclude.tex
\section{Conclusion}

In conclusion, existing Gaussian-based methods face challenges in rendering glossy objects due to specular ambiguity across camera views. In this work, we proposed a reflection-aware Gaussian Splatting method, named RGS, with a 3D foundation model prior to regularize the geometry for high-quality specular novel view synthesis. 
We first propose a cross-view shape consistency regularization strategy that exploits VGGT to impose cross-view consistency constraints for the refinement of the geometry. To better capture specular patterns, we propose a reflection-guided densification strategy that enhances specular capture. Our method achieves state-of-the-art novel view synthesis for reflective objects in terms of quantitative metrics and qualitative evaluation.